\documentclass[twocolumn]{article}

\usepackage{fullpage}
\usepackage{graphicx}
\usepackage{mathptmx}
\usepackage[utf8]{inputenc} 
\usepackage[T1]{fontenc}    
\usepackage{hyperref}       
\usepackage{url}            
\usepackage{booktabs}       
\usepackage{amsfonts}       
\usepackage{nicefrac}       
\usepackage{microtype}      
\usepackage{graphicx}
\usepackage{multirow}
\usepackage{multicol}
\usepackage{subfig}
\usepackage{amsmath}
\usepackage{lipsum}
\usepackage[dvipsnames]{xcolor}

\title{Measuring Ethics in AI with AI: \\ A Methodology and Dataset Construction}
\author{Pedro H.C. Avelar \and Rafael Baldasso Audibert \and  Anderson R. Tavares \and Luís C. Lamb}

\author{Pedro H.C. Avelar, Rafael B. Audibert, Anderson R. Tavares, Luís C. Lamb \\
              Universidade Federal do Rio Grande do Sul \\
              \{phcavelar,rbaudibert,artavares,lamb\}@inf.ufrgs.br
              }

\date{2021}

\begin{document}

\maketitle

\begin{abstract}
Recently, the use of sound measures and metrics in Artificial Intelligence has become the subject of interest of academia, government, and industry. 
Efforts towards measuring different phenomena have gained traction in the AI community, as illustrated by the publication of several influential field reports and policy documents. These metrics are designed to help decision takers to inform themselves about the fast-moving and impacting influences of key advances in Artificial Intelligence in general and Machine Learning in particular. In this paper we propose to use such newfound capabilities of AI technologies to augment our AI measuring capabilities. We do so by training a model to classify publications related to ethical issues and concerns. In our methodology we use an expert, manually curated dataset as the training set and then evaluate a large set of research papers. Finally, we highlight the implications of AI metrics, in particular their contribution towards developing trustful and fair AI-based tools and technologies. \\
\end{abstract}
Keywords: AI Ethics; AI Fairness; AI Measurement. Ethics in Computer Science.

\section{Introduction}
Recently, the use of sound measures and metrics in Artificial Intelligence (AI) has become the subject of interest of academia, government, and industry \cite{gi2018ethical,MeasureAI2020,AI_index2019,AIindex2021}. 
The widespread impact of Artificial Intelligence and Machine Learning has implied a paradigm  shift in several fields of computing research including natural language processing, machine translation, computer vision, and image recognition \cite{deepACM2021,Hinton-nature,Schmidhuber15}. Leading scientists, public leaders, and entrepreneurs including Bill Gates, Elon Musk, and the late Stephen Hawking have raised concerns about the impact of AI over every aspect of human life. These happenings   have led to increasing societal concerns about the fairness, accountability, explainability, and interpretability of AI systems and technologies \cite{burton2017ethical,ChouldechovaR20,doran2017does,FloridiCBCCDLMP18}. 
In addition, several scientific and political organizations, the United Nations, European Union, OECD, and national governments have now invested in the development of ethical guidelines and national or multilateral AI policies, regulations, and strategies \cite{bostrom-ethics,AI_index2019,UN2019}. 

One key aspect to deal with, when taking ethics into account in the development of AI systems is the use of appropriate metrics for AI ethics and policies. The Stanford's AI Index Report  \cite{AI_index2019}, in particular, addresses these and related issues. In order to better understand the societal impact of AI technologies, one has to hold a number of metrics that are useful to decision makers, including policy makers, business and technology executives, journalists, researchers and, most importantly, the general public. Educators also have an increasing responsibility as AI becomes a tool in several scientific domains, with widespread applications in every economic activity \cite{deepACM2021,burton2017ethical,ChouldechovaR20}. 

In the development of ethically bounded AI technologies, one is typically confronted with a number of challenges, including key questions on how to embed ethical principles in AI systems \cite{rossi2015safety,RossiM19}. Building intelligent agent systems, or systems that interact and work with humans in real world settings poses several challenges. 
Intelligent agents, the key components of any AI system, as argued by \cite{RossiM19} must also conciliate their subjective preferences with moral and ethical values. Thus, when specifying the ethical behaviours, boundaries, and the AI agents goals one has to seek a balance between an agent's subjective preferences and ethical boundaries, which reflect in the overall AI system behaviours \cite{mehrabi_survey_bias,RossiM19}.

These, of course, demand the development of clearly defined methodologies and metrics in the domain of AI ethics \cite{AI_index2019}. The point we make here is that - in a nutshell - AI systems must be endowed with, and subject to, clear metrics based on data driven approaches that improve the quality, fairness, explainability, and accountability of AI systems and technologies \cite{FloridiCBCCDLMP18}. These will certainly contribute to mitigate one of the most concerning characteristic of (unfortunately) more than a handful of AI systems: algorithmic biases. In turn, such biases coupled with data biases will lead systems that show undesired behaviours and prejudices \cite{parrots,crawford2013hidden,garcia2016racist}. 
In this paper, we shall contribute toward the aim of developing metrics for AI ethics. In order to do so, we propose to use the newfound capabilities of AI technologies to augment our AI-measuring capabilities. We train  an AI model to classify if a paper is ethics-related from its title and abstract descriptions. We also use expert knowledge by means of a manually curated dataset, which is used as a training set. We then  evaluate a set of papers made available in previous works, and compare the accuracy and results of our work. 

Thus, our main contributions are as follows:
\begin{enumerate}
    \item We provide a manually curated dataset of papers that present ethics-related content, which can be extracted from \url{http://arXiv.org} through their unique identifiers. We choose arXiv.org since it has become one of a \emph{de facto} open repository for most AI papers that will  be published at mainstream AI conferences and venues.
    \item We provide a trained model to evaluate whether a paper is ethics-related or not, which can be used as a tool by ethicists (and other scientists) to help them do a primary analysis of the growing amount of AI and CS-related papers that is useful to their research.
    \item We evaluate our results by comparing them with an earlier methodology for measuring the role of ethics in AI research. We do so by running our classifier through the same dataset of papers abstracts from a previous study \cite{prates2018ethics}, and we run their methodology on our test set to compare the results.
\end{enumerate}

The remainder of the paper is organised as follows. In Section \ref{relatedwork} we highlight recent research which tackles key questions in developing metrics for AI ethics. Next, we describe a dataset for identifying ethics in AI research which shall be used in our methods and experiments. 
 We then introduce a new AI-Based index that outputs the recent impact of Ethics in key AI conferences and venues.
Finally, we conclude and point out directions for further research. 

\section{Background and Related Work} \label{relatedwork}
Recent efforts towards measuring different phenomena in the AI community have gained traction, as illustrated e.g. through the Stanford Human-Centered AI Institute's AI Index efforts \cite{AI_index2019,AIindex2021}. These AI metrics are designed to help policy-makers and researchers to inform themselves about this fast-moving and impacting field. In this paper, we propose to use the newfound capabilities of AI technologies to augment our measuring capability, training a model to classify if a paper is ethics-related from the information provided in its abstract contents.

Ethical concerns on the implications of the data-driven scientific paradigm, reinforced by the prominence of AI technologies and methods,  has also raised a number of societal concerns. For instance, Green \cite{green2020data} has recently urged those who apply data-driven artificial intelligence and machine learning to social and political contexts to acknowledge the impacts of their products and take a more firm stance as policy actors. They discuss three of the main excuses used by engineers to avoid taking stances with regards to how their products are used: the first, that one is "simply an engineer" and does not dictate how the technology they produce will be used; second, that it is not the data scientist's "job" to take a political stance and that remaining neutral during the development is the best course of action; third, that perfectly managing a technology's impact is unfeasible and that this should not be an impediment to create new products that improve society incrementally. Green \cite{green2020data} opposes these arguments defending a apolitical stance and then proceeds to discuss one possible path to incorporating principles into data science to strive for social justice. 

In \cite{garzcarek2019approaching} the authors call for data scientists to recognise themselves as a group and discusses issues raised as early as 2017 by France's Commission Nationale de l'Informatique et des Libertés \cite{cnil2017ethics} regarding the technologies produced by data scientists. They  further discuss the various already-available ethical frameworks that data scientists can use as an azimuth, most of which had been updated in 2018, such as the codes of conduct from the American Statistical Association \cite{asa2018ethical}, Association for Computing Machinery \cite{acm2018ethical} and German Informatics Society \cite{gi2018ethical}, while also stating that uniting data scientists does not imply into forming a new society and build a code of conduct from the ground up, but rather that these already existing codes could have Data-Science-specific guidelines added to them. Further,  \cite{wilk2019teaching} raises the topics brought about in such codes of conduct and proposes a course to help professionals in the area of AI, as well as regulators and policy makers, to come to terms with the many and shifting ethical issues brought about by the AI paradigm change implications, as well as the ensuing legal and regulatory debates and constraints.  It is in this social and scientific context that our work is situated: AI has now gone well beyond the realm of technology and has reached ubiquitous use; therefore, measuring the social consequences of AI use is paramount to guarantee that its tools and technologies will have positive impacts in human life. 

\subsection*{A Note on Dataset Classification} 
In order to carry out our investigation, we use a manually curated dataset as a training set and evaluate a set of papers made available in previous work, comparing our results with them. We chose not to use crowdsourcing or mechanical turk to classify our data. This is due to both ethical and technical reasons (see e.g. \cite{ratcliff2021data} for a deeper discussion on using such techniques do classify specific kinds of data). Further, the recent work of \cite{shmueli2021beyond} analyses several ethical implications of using mechanical turk in natural language processing, AI and data analyses. Considering ethical implications and the fact that the field of AI ethics and fairness is novel, we opted to use expert knowledge to classify the papers in our dataset. We claim that to better understand whether certain a certain piece of research is qualified for our analyses, one has to refer to expert knowledge. As we are analysing a body of technical work that demands skilled knowledge of AI, an expert curated dataset allows in principle a  better understanding if a certain piece of research is related to ethics. In addition, an expert can possibly better evaluate if 
the AI research, tool, method, and technology described in a paper might have ethical or social consequences over third parties. 

Along these lines, we highlight that an earlier study provided a metric for ethics in AI research \cite{prates2018ethics} that has been used in the last two Stanford's AI Index editions \cite{AI_index2019,AIindex2021}. In such work, the authors were able to measure and analyse and use of ethics-related terms in flagship AI conferences and journals over the last fifty years. In a nutshell, their results show that, although AI was seen as a field that potentially impacted human life in the last decades, technical research papers typically do not explicitly analyse the ethical implications of their research results, tools, technologies, and achievements. One has also to mention that our work contributes towards disseminating a culture of principles 
Principled AI research, as defended recently by researchers, corporations, public, governments, and social organizations \cite{fjeld2020principled,OECD2019,UN2019}. 

\section{On Building a Dataset for Measuring Ethics in AI}\label{sec:dataset}

We collected a total of 238806 papers from Arxiv, ranging from 1989-12-31 up until 2019-10-23. These papers' metadata contained a list field ``category'', which unfiltered amounted to about 10441 categories. Filtering for only those papers which had an ``abs'' identified, there were 9839 categories. In the end, we filtered papers which had either ``cs.cy'' or ``Computers and Society'' (to filter for ethics-related papers) and had any of 
``cs.AI", ``Artificial Intelligence", ``cs.CL", ``Computation and Language", ``cs.CV", ``Computer Vision", ``Pattern Recognition", ``cs.MA", ``Multiagent Systems", ``cs.LG", ``Learning", ``cs.NE", ``Neural and Evolutionary Computing", ``stat.ML", ``Machine Learning" (to filter for AI-related papers) inside one of their category identifiers. Filtering for these we were left with 1425 papers to be annotated as related to the field or had contents associated with AI ethics. 

With this subset of 1425 papers, the authors then proceeded to manually annotate and curate 200 of the papers on whether they were AI-related or ethics-related, based solely on their titles and abstracts. The final decision was done via a majority vote, where the paper would take on the label that the majority of the researchers assigned to it. The vast majority of papers were already in the AI category due to how the data was collected, however, of the 200 annotated papers there were only 54 papers which were considered to be about ethics, the rest 146 being annotated as not ethics-related.

\subsection{Active Learning}

To increase the number of papers available in the dataset we used active learning \cite{lewis1994uncertainty,settles2009active} to augment the dataset with machine-labelled examples. To do so, we did two rounds of machine labelling, with the first also serving as a model validation step. To keep closer to the index produced with \cite{prates2018ethics}, we abstained from using complex NLP models such as transformers or recurrent neural networks, depending on simpler models depending instead on a Bag-of-Words or Term-Frequency representation.

We evaluated hyperparameter combinations for models using a 4-fold cross validation, testing models such as Logistic Regression, Adaptive Boosting, Gradient Boosting, Decision Trees, Random Forests, and Multilayer Preceptrons, the combinations which were tested can be seen in Table~\ref{tab:hyperparams}.

\begin{table}[hptb]
    \small
    \centering
    \begin{tabular}{ccc}
    \toprule
        Model & Hyperparameter & Options \\
    \midrule
        \multirow{2}{*}{LR}         & Inv. Reg. Strength       & 0.25, 0.5, 1, 2, 4 \\ 
                                    & Regularisation           & $l1$, $l2$ \\ \midrule
        \multirow{2}{*}{AB}   & Num. of Estimators       & 8, 32, 128, 512 \\ 
                                    & Learning Rate            & 0.125, 0.25, 0.5, 1 \\ \midrule
        \multirow{2}{*}{GB}  & Num. of Estimators       & 8, 32, 128, 512 \\
                                    & Max. Tree Depth          & 1, 2 , 4, 8 \\ \midrule
        \multirow{2}{*}{RF}         & Num. of Estimators       & 8, 32, 128, 512 \\
                                    & Max. Tree Depth          & 1, 2 , 4, 8 \\ \midrule
        \multirow{3}{*}{DT}         & Optimisation Criterion   & Gini, Entropy \\
                                    & Splitting Method         & Best, Random \\
                                    & Max. Tree Depth          & 1, 2 , 4, 8 \\
        \multirow{4}{*}{MLP}        & Activation function      & TanH, ReLU \\ \midrule
                                    & Learning Rate            & $10^{-3}$, $10^{-4}$, $10^{-5}$ \\
                                    & Learning Rate Technique  & Adaptative, Constant \\
                                    & Max. Training Iterations & 32, 128, 512 \\
    \bottomrule
    \end{tabular}
    \caption{Models and Hyperparameters used in the experiments. LR stands for Logistic Regression, AB for Adaptive Boosting, GB for Gradient Boosting, RF for Random Forest, DT for Decision Tree, MLP for Multilayer Perceptron. For the MLP model we used the Adam optimiser.}
    \label{tab:hyperparams}
\end{table}

Due to the highly imbalanced nature of the dataset, we used oversampling to create a balanced version of the dataset. We avoided using oversampling techniques that generated synthetic value such as SMOTE and ADASYN since the input values do nor represent numeric values, and we did not want to perform generative sampling without a generative model that worked on the original data format. We also tested the model without handling the imbalance on the dataset, but it produced models with a significantly lower ROC-AUC score, and a higher tendency to reject papers as not being AI-related.

Given the initial model choices, we ended up with using a random forest classifier with 512 estimators and a maximum depth of 8. We used L1 norm on the term frequencies and used IDF scaling. Given this, we proceeded with two rounds of model training and classification. On the first round the best classifier obtained an average ROC-AUC score of 0.98 on the 4-fold cross validation step, leaving us with the final model being a Random Forest with 512 estimators and a maximum tree depth of 8. The model when trained on the entirety of the 200 samples misclassified two ethics-related paper as not being on ethics. Furthermore, one of the human classifiers classified 300 more papers with which the model's results were compared, where it agreed 83.33\% with the human labeler.

After this step, the human classifiers classified another 79 papers where the model was unsure of its predictions \cite{lewis1994uncertainty}, where we considered it sure of its prediction if the label probability was lower than 1/3 or higher than 2/3. Given this new labelled set, containing 83 ethics-related papers on a total of 543 labelled samples, we did a 4-fold cross validation step on model with the same hyperparameter settings, obtaining an average ROC-AUC score of .99. We then re-trained the model on the all the labelled data and labelled the rest of the dataset with the second model's label whenever a human-annotated label was not available. These entries are specified in the dataset we will make available so that future users can know which entries are machine-labelled if they wish to avoid them.

\subsection{Dataset Analysis}\label{ssec:dsetanalysis}

The final dataset we will make available with this paper has 290 hand-labelled examples, with the other 1136 being machine-labelled examples using a bag-of-words interpretation of the document and a random first classifier. Of these, 21.61\% were considered to be ethics-related.

Finally, we can use the methodology proposed in \cite{prates2018ethics} and used in the AI-Index \cite{AI_index2019} to assess their technique for identifying ethics-related material from abstracts and titles in our dataset. First testing on the human-labelled sampled their model achieved low scores, having a ROC AUC score of 0.68, with 68\% precision and 45\% recall, both severely underestimating the number of ethics-related papers while also producing some false positives.

When ran on the entirety of the dataset, the model had similar results, woth a ROC AUC score of 0.86, 54\% precision and 48\% recall. This result \emph{further motivates our approach}, which tries to build a more robust method for pinpointing where the discussion about ethics in AI is, using machine learning models to help identify these papers.

\section{The Construction of the AI-Based AI-Index} \label{sec:methodology}

In order to construct an AI-based index, We shall use the same datasets as made available by the authors of \cite{prates2018ethics} to perform experiments on generating an AI-based AI-Index. The available data were paper abstracts from flagship conferences (including e.g. AAAI and NeurIPS) as well as paper titles from flagship AI and robotics conferences and journals, selected by the authors. Such data shall then be analysed through a model trained on the aforementioned data.

\subsection{A Logistic Regression Model} \label{subsec:logistic}

As a first study on how an alternative AI-based AI-Index model would work, we trained a logistic regression model with $l1$ normalisation. Due to how $l1$ normalisation works, the model would have weights lying close to the unit square, serving either as a positive input, which would serve to classify a paper as being on ethics-related topics, or a negative input, which would harm a paper's likelihood to be classified as being on an ethics-related topic. These keywords can give us insight on the composition of the dataset as well as being able to assess likely failings that a more complex AI-based model might associate with these papers.

The list of keywords used as ethics-related on \cite{prates2018ethics} were: \emph{accountability, accountable, employment, ethic, ethical, ethics, fool, fooled, fooling, humane, humanity, law, machine bias, moral, morality, privacy, racism, racist, responsibility, rights, secure, security, sentience, sentient, society, sustainability, unemployment, workforce}. 
However, since we have been using lemmatisation in our models so far, through which the list of ethic-related lemmas would be: \emph{accountability, accountable, employment, ethic, ethical, fool, humane, humanity, law, machine bias, moral, morality, privacy, racism, racist, responsibility, right, secure, security, sentience, sentient, society, sustainability, unemployment, workforce}.

Thus, training a Logistic Regression model as specified above on our dataset, we obtained a model that weighted positively the following lemmatised keywords: \emph{ai, bias, discrimination, ethical, fair, fairness, how, human, machine, may, social, these, trust. And weighted negatively the following lemmas: by, datum,  information, method, model, network, propose, student, time, use}. The model had an intercept of -0.59, even though we used oversampling to balance the dataset.

The list of keywords learned by the model seemed to have a similar vein to that presented by \cite{prates2018ethics}, however it was not without its failings. First of all, some reasonably generic keywords were introduced in the model, such as "ai", "how", "human", "machine", "may" and "these", with the first and fourth ones most likely being added due to the bias our dataset has towards papers that talk about ethics in AI. An interesting note to be made, however, is that the model seemed to balance some of these generic keywords with other generic keywords on the negative part, such as \emph{by, datum, method, model, network, propose} -- all of which could be interpreted as pointing more towards an AI/ML model and further away from a paper discussing ethics in AI.

Another large failing from the AI-generated list is that it lacks keywords representing some important topics, such as AI accountability, the impact of AI in employment, AI's reinforcement of biases with regards to race, the relation of AI with law, and questions about its security (although one may argue that this last topic might've been slightly touched by the "trust" keyword). This shows a great gap that still needs to be bridged with regards to building a dataset that encompasses all of these topics.

\subsection{Re-Analysing the AI-Index} \label{subsec:reanalysing}

The analyses presented in \cite{prates2018ethics} and used in the last two Stanford's AI-Index Reports \cite{AI_index2019,AIindex2021} analysed the frequency of keywords considered as ethics-related in flagship AI and robotics conferences as well as top AI and Robotics journals. Here we perform the analyses along the lines of the one done by \cite{prates2018ethics}, but using the random forest model trained with the dataset we will make available. The first part of this study follows closely what we have done so far; their study analysed the frequency of these keywords in paper abstracts for two conferences, namely AAAI and NeurIPS. The second part, however only used paper titles.

To improve on this issue, for this part of the analysis we train our model with a dataset containing both titles and abstracts, so the model has to learn to predict if a paper is ethics-related both using its abstract and  using only its title. From a preliminary study training on paper abstracts and testing on paper titles we noticed a significant drop in the model's performance, and thus decided to continue using a model trained on both alternatives.

\begin{figure*}[tbhp]
    \centering
    \subfloat[]{
        \includegraphics[width=.50\linewidth]{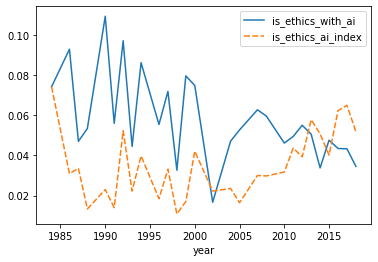}
        \label{fig:abs:aaai}
    }
    \subfloat[]{
        \includegraphics[width=.50\linewidth]{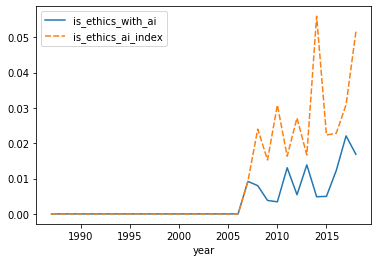}
        \label{fig:abs:nips}
    }
    \caption{Number of documents classified as ethics related in AAAI (Subfigure~\ref{fig:abs:aaai}) and NeurIPS (Subfigure~\ref{fig:abs:nips}) abstracts by our model "is\_ethics\_with\_ai" and by following \cite{prates2018ethics} "is\_ethics\_ai\_index".}
    \label{fig:abs}
\end{figure*}

As we can see in Figure \ref{fig:abs}, the model we produced disagreed slightly with the work of \cite{prates2018ethics}, that is a keyword-based classification, even though it seemed to maintain some of the peaks in Subfigure~\ref{fig:abs:aaai} in 1990, 1991, 1994, 1997, and the year 2000; on the other years, the model seemed to estimate a larger amount of ethics-related papers than the keyword-based model, and showed a decreasing tendency in the last years.

In Subfigure~\ref{fig:abs:nips}, however, the model estimated a much smaller amount of papers as being ethics-related, than what the keyword-based model estimated. Nonetheless, there still seemed to be an updward trend in more recent conferences, and it predicted a dip in the years of 2010, 2012 and 2014, where the keyword-based model estimated a peak instead.

\begin{figure*}[tbhp]
    \centering
    \subfloat[]{
        \includegraphics[width=.5\linewidth]{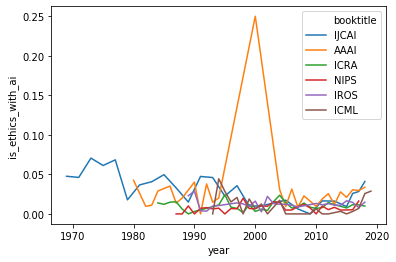}
        \label{fig:conferences:ai}
    }
    \subfloat[]{
        \includegraphics[width=.5\linewidth]{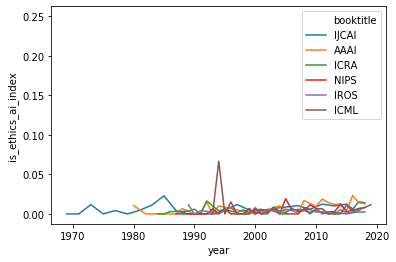}
        \label{fig:conferences:index}
    }
    \caption{Number of documents classified as ethics related in the selected flagship conferences' titles. The graph on \ref{fig:conferences:ai} is our method and on \ref{fig:conferences:index} is using the keywords presented in \cite{prates2018ethics}}
    \label{fig:conferences}
\end{figure*}

\begin{figure*}[bthp]
    \centering
    \subfloat[]{
        \includegraphics[width=.5\linewidth]{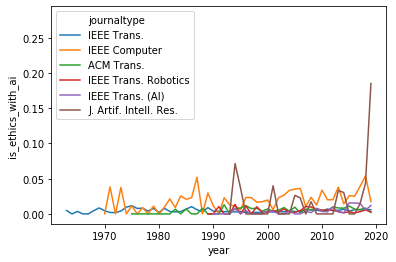}
        \label{fig:journals:ai}
    }
    \subfloat[]{
        \includegraphics[width=.5\linewidth]{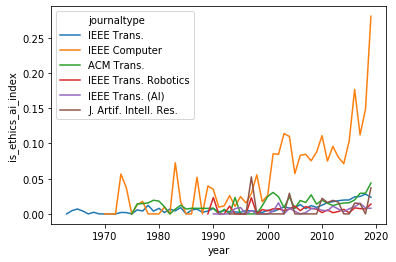}
        \label{fig:journals:index}
    }
    \caption{Number of documents classified as ethics related in the selected journal venues' titles. The graph on \ref{fig:conferences:ai} is our method and on \ref{fig:conferences:index} is using the keywords presented in \cite{prates2018ethics}}
    \label{fig:journals}
\end{figure*}

In Figures~\ref{fig:conferences} and \ref{fig:journals}, however, is where we see the biggest disagreements between the two approaches, with both having very different scales.

In the conference titles (Fig.~\ref{fig:conferences}) we can see that the AI-based model seemed to predict more papers as being ethics-related in older conferences, where the keyword-based model predicted very few. Another big difference is that the AI-based model has a very significant peak in the AAAI conference of the year 2000, where the keyword-based one only had a much smaller peak in 1994. The 2000's peak is caused due to the fact that the dataset only had 4 paper titles for AAAI in the year 2000, among which the title "Artificial Intelligence-Based Computer Modeling Tools for Controlling Slag Foaming in Electric Furnaces." was (we believe, wrongly) classified as being ethics-related.

Now, in the journal titles (Fig.~\ref{fig:journals}) we see that the keyword-based model detects and upward trend in for the journal "IEEE Computer", where the AI-based model only predicts a peak in the last year for the Journal of Artificial Intelligence Research, seemingly having a stable state in the other years and journals. Looking closely at the titles classified as ethics-related by either model in the last year (2019) for both these journals we saw that many of the places where the models disagreed were on titles we believed not to be ethics-related, with only 2 of the papers in IEEE Computer being unanimously classified as such, and only 5 in total in both journals having a majority of us considering them as ethics-related. Also, the spike present in the JAIR journal for the year 2019 was mostly because of how few papers were available in the dataset used in \cite{prates2018ethics}.

We also present, like in \cite{prates2018ethics}, tables with the percentage of papers that are considered to be AI-related by the algorithm, which serves a double purpose of being another way to identify a conference's or journal's ethics in  AI participation, as well as an anomaly detection for the provided algorithm.

For example, in Table~\ref{tab:conferences} we can see the proportion of papers which the model considered to be ethics-related. From looking at these numbers we can quickly detect some anomalies in the model, for example the model classified 1 of the 4 papers present in the dataset for AAAI 2000 as being ethics-related, which we can easily see to not be true by looking at the paper itself, whose name was "Artificial Intelligence-Based Computer Modeling Tools for Controlling Slag Foaming in Electric Furnaces" -- clearly not ethics related. However, this over-reporting is less self-evident in venues which have more occurrences, since the model still tends to predict more papers as non-ethics related. In Table~\ref{tab:journals}, we can see a similar table for the analysed journal venues.

\begin{table}[hptb]
    \small
    \centering
    \begin{tabular}{ccccccc}
    \toprule
        Year & AAAI & ICML & ICRA & IJCAI & IROS & NeurIPS\\
    \midrule
1980 & 0.043 & - & - & - & - & -\\
1981 & - & - & - & 0.037 & - & -\\
1982 & 0.010 & - & - & - & - & -\\
1983 & 0.011 & - & - & 0.041 & - & -\\
1984 & 0.029 & - & 0.014 & - & - & -\\
1985 & - & - & 0.012 & 0.050 & - & -\\
1986 & 0.035 & - & 0.015 & - & - & -\\
1987 & 0.014 & - & 0.015 & 0.033 & - & 0.000\\
1988 & 0.020 & - & 0.006 & - & - & 0.000\\
1989 & - & - & 0.000 & 0.015 & 0.023 & 0.010\\
1990 & 0.040 & - & 0.003 & - & 0.029 & 0.000\\
1991 & 0.000 & - & 0.005 & 0.047 & 0.004 & 0.007\\
1992 & 0.038 & - & 0.007 & - & 0.003 & 0.008\\
1993 & 0.015 & 0.000 & - & 0.046 & 0.010 & 0.006\\
1994 & 0.020 & 0.044 & 0.011 & - & 0.011 & 0.007\\
1995 & - & 0.029 & 0.024 & 0.023 & - & 0.000\\
1996 & - & 0.015 & 0.009 & - & 0.013 & 0.007\\
1997 & - & 0.021 & 0.007 & 0.036 & 0.014 & 0.006\\
1998 & - & 0.000 & 0.002 & - & 0.013 & 0.020\\
1999 & - & 0.019 & 0.012 & 0.010 & 0.010 & 0.007\\
2000 & 0.250 & 0.007 & 0.003 & - & 0.016 & 0.007\\
2001 & - & 0.013 & 0.006 & 0.010 & 0.003 & 0.010\\
2002 & - & 0.000 & 0.004 & - & 0.022 & 0.010\\
2003 & - & 0.009 & 0.016 & 0.014 & 0.013 & 0.015\\
2004 & 0.031 & 0.017 & 0.024 & - & 0.012 & 0.015\\
2005 & 0.009 & 0.000 & 0.014 & 0.017 & 0.012 & 0.005\\
2006 & 0.031 & 0.000 & 0.007 & - & 0.011 & 0.005\\
2007 & 0.008 & 0.000 & 0.009 & 0.006 & 0.009 & 0.009\\
2008 & 0.023 & 0.000 & 0.014 & - & 0.010 & 0.012\\
2009 & - & 0.000 & 0.008 & 0.000 & 0.012 & 0.008\\
2010 & 0.010 & 0.006 & 0.007 & - & 0.013 & 0.000\\
2011 & 0.019 & 0.000 & 0.008 & 0.016 & 0.011 & 0.010\\
2012 & 0.026 & 0.000 & 0.014 & - & 0.013 & 0.005\\
2013 & 0.012 & - & 0.012 & 0.016 & 0.011 & 0.008\\
2014 & 0.028 & 0.003 & 0.010 & - & 0.011 & 0.005\\
2015 & 0.021 & 0.000 & 0.008 & 0.009 & 0.017 & 0.005\\
2016 & 0.031 & 0.003 & 0.012 & 0.026 & 0.015 & 0.005\\
2017 & 0.030 & 0.007 & 0.011 & 0.028 & 0.008 & 0.016\\
    \bottomrule
    \end{tabular}
    \caption{Ratio of papers that were considered by the model to be ethics-related topic on the four AI and two Robotics conferences.}
    \label{tab:conferences}
\end{table}

We also take 5 random examples from the conference titles and another 5 for the journal titles to see on what cases one model or the other might fail in discerning the values correctly. In the following paragraphs we will present the paper title between double quotes and an indication of the proportion of authors that thought the paper was ethics-related. So, if one third of the authors agreed that the paper was ethics-related, a paper would appear as "title" (1/3). If none of the authors thought the paper to be ethics, related it would only appear as "title".

In the journal titles dataset, the keyword-based method predicted that the following 4 titles were titles of ethics-related papers: "Secure and Efficient Handover Authentication Based on Bilinear Pairing Functions.", "When Does Relay Transmission Give a More Secure Connection in Wireless Ad Hoc Networks?", "Performance of the biased square-law sequential detector in the absence of signal.", and "Secure program partitioning.". Of these 4, we believe that none of them could be considered ethics-related by looking only at the title. However, the title only the Random Forest classified as ethics-related -- "Virtual Character Facial Expressions Influence Human Brain and Facial EMG Activity in a Decision-Making Game." (1/3), might be considered as ethics-related by some, although most of us who reviewed it believed it not to be so.

In the conference titles, we had that the AI-based method judged the titles "Human-machine skill transfer extended by a scaffolding framework.", "Enhanced manipulator's safety with artificial pneumatic muscle.", and "Logic Programing in Artificial Intelligence." were considered to be ethics-related, while the keyword-based method classified as such the following two titles: "Coherence of Laws." (2/3), "Efficient Methods for Privacy Preserving Face Detection." (2/3). Here, the lack of relevant papers discussing law and privacy in our training set -- made apparent by the lack of these keywords as discussed in Subsection~\ref{subsec:logistic} -- is clear, with the two titles that our method did not classify as ethics-related, where  the keyword-based model did, were exactly about these topics.

One important aspect to note about the more in-depth analysis of the disagreements between the two models is that in very few of the classifications there was a unanimous agreement about the paper being ethics-related, so even when the models theoretically should have voted in them being ethics-related, an agreement between expert human labellers was not reached. The only cases where an agreement was reached were in titles such as "Codes of Ethics in a Post-Truth World." and "Algorithms: Law and Regulation.", one of which was captured by the catch-all "ethics" keyword and the other touched aspects of law in algorithms, but did not necessarily inform us in the title whether it was about artificial intelligence or not.

\begin{table}[hptb]
    \small
    \centering
    \begin{tabular}{ccccccc}
    \toprule
        Year & CACM & C  & T. & TAI & TR & JAIR\\
    \midrule
    1980 & 0.000 & 0.000 & 0.002 & - & - & -\\
    1981 & 0.000 & 0.010 & 0.008 & - & - & -\\
    1982 & 0.000 & 0.021 & 0.003 & - & - & -\\
    1983 & 0.007 & 0.009 & 0.003 & - & - & -\\
    1984 & 0.000 & 0.025 & 0.003 & - & - & -\\
    1985 & 0.007 & 0.021 & 0.007 & - & - & -\\
    1986 & 0.000 & 0.023 & 0.010 & - & - & -\\
    1987 & 0.000 & 0.052 & 0.006 & - & - & -\\
    1988 & 0.008 & 0.000 & 0.003 & - & - & -\\
    1989 & 0.000 & 0.030 & 0.009 & - & 0.000 & -\\
    1990 & 0.000 & 0.012 & 0.004 & 0.000 & 0.000 & -\\
    1991 & 0.000 & 0.000 & 0.004 & 0.000 & 0.010 & -\\
    1992 & 0.013 & 0.023 & 0.003 & 0.000 & 0.000 & -\\
    1993 & 0.000 & 0.013 & 0.003 & 0.000 & 0.000 & 0.000\\
    1994 & 0.008 & 0.011 & 0.003 & 0.008 & 0.014 & 0.071\\
    1995 & 0.008 & 0.006 & 0.003 & 0.000 & 0.000 & 0.037\\
    1996 & 0.012 & 0.023 & 0.002 & 0.005 & 0.011 & 0.000\\
    1997 & 0.008 & 0.023 & 0.002 & 0.000 & 0.000 & 0.000\\
    1998 & 0.008 & 0.017 & 0.002 & 0.000 & 0.010 & 0.000\\
    1999 & 0.004 & 0.017 & 0.002 & 0.000 & 0.002 & 0.000\\
    2000 & 0.007 & 0.020 & 0.004 & 0.004 & 0.002 & 0.000\\
    2001 & 0.004 & 0.006 & 0.003 & 0.004 & 0.005 & 0.040\\
    2002 & 0.005 & 0.023 & 0.004 & 0.000 & 0.002 & 0.000\\
    2003 & 0.009 & 0.027 & 0.003 & 0.000 & 0.007 & 0.000\\
    2004 & 0.004 & 0.033 & 0.005 & 0.004 & 0.002 & 0.000\\
    2005 & 0.010 & 0.035 & 0.003 & 0.000 & 0.002 & 0.026\\
    2006 & 0.002 & 0.036 & 0.003 & 0.000 & 0.005 & 0.023\\
    2007 & 0.007 & 0.011 & 0.006 & 0.004 & 0.010 & 0.000\\
    2008 & 0.006 & 0.024 & 0.006 & 0.002 & 0.010 & 0.017\\
    2009 & 0.008 & 0.013 & 0.005 & 0.008 & 0.007 & 0.000\\
    2010 & 0.005 & 0.034 & 0.004 & 0.003 & 0.005 & 0.000\\
    2011 & 0.005 & 0.020 & 0.007 & 0.003 & 0.006 & 0.000\\
    2012 & 0.010 & 0.020 & 0.004 & 0.010 & 0.005 & 0.000\\
    2013 & 0.008 & 0.038 & 0.005 & 0.003 & 0.003 & 0.033\\
    2014 & 0.008 & 0.014 & 0.007 & 0.002 & 0.002 & 0.030\\
    2015 & 0.011 & 0.026 & 0.004 & 0.016 & 0.003 & 0.000\\
    2016 & 0.007 & 0.025 & 0.006 & 0.015 & 0.001 & 0.000\\
    2017 & 0.006 & 0.039 & 0.006 & 0.015 & 0.003 & 0.015\\
    \bottomrule
    \end{tabular}
    \caption{Ratio of papers that were considered by the model to be on ethics-related topics on the analysed Journal Venues; respectively, [CACM] Communications of the ACM, [C] IEEE Computer, [T] IEEE Transactions, [TAI] IEEE Transactions (AI), [TR] IEEE Transactions on Robotics, and [JAIR] Journal of Artificial Intelligence Research.}
    \label{tab:journals}
\end{table}

\section{Limitations}

We acknowledge here that our model is not without its limitations. First of all, the use of active learning as a strategy to classify the papers was done so due to a lack of resources. If the ethics community could in turn produce a large amount of reliably classified papers, one could certainly build a probably more robust model than we have achieved. Nonetheless, our model is a step in the right direction of using a data-based solution to provide such information, and the dataset we produced is undoubtedly a contribution that can be built upon to improve this area of research.

Another issue of our work is that what defines a paper as being ``ethics-related''  is based on our experience in the subject. Our working group did not let each others' classification influence each other (we did a double-blind classification), and probably even if we did discuss each of the controversial classifications, it is unlikely that we could reach an agreement in all cases. We believe, again, that a large (however, ethic) paper  classification could help iron-out these discrepancies and provide classifications that are aligned with what is ``commonsense'' in the community.

Another issue one might have with our model is that we used a bag-of-words representation, and a logistic regression model to classify this paper. We justify this as to keep the model as interpretable as possible while being aligned with previous work regarding the use of a bag-of-words to classify a paper as being ethics-related or not \cite{prates2018ethics}.

\section{Discussion}\label{sec:discussion}

In this paper we provided an AI-powered tool for classifying research papers as being ethics-related from its own abstract. We provide a first use of this for measuring the AI community's engagement with ethics-related research in the main tracks of flagship venues  proceedings. As a consequence, we were able to identify both its characteristics and the keyword-based model's failings, providing some insight on these disagreements with the keyword-based model. This allows us to showing that many of the papers previously reported as being ethics-related might actually be wrongly classified as so, and that our model, although imperfect due to the limited scale of labelled data used for its training, helps alleviate this in some cases.

However, one must be aware that the proposed techniques can be improved. First, the data regime used to train this model is thinner than the one common to most machine learning approaches. Second, natural language models are known to exhibit biases contained in the data they have been trained, so one should be mindful that the model we provide here is not without its flaws, and should still be improved in order to classify correctly all the papers provided to it. Especially, we pointed out the lack of papers discussing law and race in our training set, which might hinder the performance of our model in detecting papers on these topics, but this, in turn, is a feature of most AI flagship venues so fat, which typically have published a limited number of research papers on the social implications of AI tools and methods. 
\section*{Acknowledgements}
This work was supported in part by the Brazilian National Research Council CNPq and by the CAPES Foundation, Finance Code 001. \\ {Data from the experiments described in this paper will be made available at a future date.}

\bibliographystyle{plain}
\bibliography{ethicswithai}
\end{document}